\documentclass[journal,twoside,web]{ieeecolor}
\usepackage{generic}
\usepackage{cite}
\usepackage{amsmath,amssymb,amsfonts}
\usepackage{algorithmic}
\usepackage{graphicx}
\usepackage{algorithm,algorithmic}
\usepackage{hyperref}
\hypersetup{hidelinks=true}
\usepackage{textcomp}
\usepackage{multirow}
\usepackage{booktabs}
\usepackage{array}
\usepackage{threeparttable}
\usepackage{subcaption}
\def\BibTeX{{\rm B\kern-.05em{\sc i\kern-.025em b}\kern-.08em
    T\kern-.1667em\lower.7ex\hbox{E}\kern-.125emX}}
\begin{document}
\title{Benchmarking Foundation Models with Multimodal Public Electronic Health Records}
\author{Kunyu Yu, Rui Yang, Jingchi Liao, Siqi Li, Huitao Li, Irene Li, Yifan Peng, Rishikesan Kamaleswaran, and Nan Liu, \IEEEmembership{Senior Member, IEEE}
\thanks{This work was supported by the Duke-NUS Signature Research Programme funded by the Ministry of Health, Singapore. \textit{(Corresponding author: Nan Liu.)} }
\thanks{Kunyu Yu, Rui Yang, Jingchi Liao, Siqi Li and Huitao Li is with Centre for Quantitative Medicine and Duke-NUS AI + Medical Science Initiative, Duke-NUS Medical School, Singapore, Singapore. }
\thanks{Irene Li is with Graduate School of Engineering, The University of Tokyo, Tokyo, Japan.}
\thanks{Yifan Peng is with the Department of Population Health Sciences, Weill Cornell Medicine, New York, NY, USA.}
\thanks{Rishikesan Kamaleswaran is with the Department of Surgery, Duke University School of Medicine, Durham, NC, USA.}
\thanks{Nan Liu is with Centre for Quantitative Medicine, Duke-NUS AI + Medical Science Initiative and Programme in Health Services and Systems Research, Duke-NUS Medical School and NUS Artificial Intelligence Institute, National University of Singapore, Singapore, Singapore (e-mail: liu.nan@duke-nus.edu.sg).}}

\maketitle

\begin{abstract}
Foundation models have emerged as a powerful approach for processing electronic health records (EHRs), offering flexibility to handle diverse medical data modalities. In this study, we present a comprehensive benchmark that evaluates the performance, fairness, and interpretability of foundation models, both as unimodal encoders and as multimodal learners, using the publicly available MIMIC-IV database. To support consistent and reproducible evaluation, we developed a standardized data processing pipeline that harmonizes heterogeneous clinical records into an analysis-ready format. We systematically compared eight foundation models, encompassing both unimodal and multimodal models, as well as domain-specific and general-purpose variants. Our findings demonstrate that incorporating multiple data modalities leads to consistent improvements in predictive performance without introducing additional bias. While domain-specific fine-tuning offers a cost-effective solution for unimodal foundation models, this effectiveness does not translate well to multimodal scenarios. Additionally, our experiments reveal limited task generalizability in current large vision-language models (LVLMs), emphasizing the need for more versatile and robust medical LVLMs. Through this benchmark, we aim to support the development of effective and trustworthy multimodal artificial intelligence (AI) systems for real-world clinical applications. Our code is available at \url{https://github.com/nliulab/MIMIC-Multimodal}.
\end{abstract}

\begin{IEEEkeywords}
Artificial intelligence, Foundation models, Large vision-language models, Multimodal electronic health records, Trustworthy AI
\end{IEEEkeywords}

\section{Introduction}
\label{sec:introduction}
\IEEEPARstart{E}{lectronic} health records (EHRs) constitute the fundamental data infrastructure supporting modern healthcare systems by systematically collecting patient information to facilitate effective clinical management and healthcare delivery~\cite{tang2024harnessing}. EHRs encompass longitudinal, heterogeneous data from diverse medical settings, including demographics, vital signs, diagnoses, medications, imaging, and clinical notes, providing a comprehensive insight into patients’ health histories~\cite{abul2019personalized,jensen2012mining}. The widespread adoption of EHR systems has not only enhanced healthcare quality but also unlocked vast data resources, enabling the identification of clinical patterns across diverse patient populations and generating more reliable and generalizable real-world evidence~\cite{tang2024harnessing}.

However, privacy concerns surrounding sensitive clinical data pose significant challenges for cross-institutional data sharing and collaborative research~\cite{de2023guide}. To mitigate these challenges, large-scale public EHR databases, such as MIMIC-IV~\cite{johnson2023mimic}, have been established. By implementing rigorous de-identification strategies and credentialed access protocols, these databases facilitate more democratized and secure access to extensive clinical data, thus fostering the advancement of data-driven medical AI solutions~\cite{ke2024comparing}. Despite these benefits, the curation and harmonization of highly heterogeneous datasets remain critical technical challenges, mainly due to complex data linkage and inconsistent data quality~\cite{acosta2022multimodal}.

With the rapid advancement in artificial intelligence (AI), foundation models have arisen as a transformative approach, capable of efficiently parsing EHR data to support clinical decision-making~\cite{moor2023foundation,tu2024towards}. Unlike traditional methods, foundation models leverage large-scale data pre-training to build representations that can subsequently be adapted,  through fine-tuning, to various downstream tasks~\cite{bommasani2021opportunities}. Owing to the inherently multimodal nature of clinical decision-making, there has been substantial interest in developing foundation models that integrate diverse, heterogeneous EHR data sources~\cite{liu2022multimodal,li2024multimodal}. 

Achieving effective multimodal learning requires capturing not only modality-specific information but also the complex relationships across different modalities~\cite{he2024foundation}. Foundation models are well-positioned to address both challenges by learning rich representations within each modality and seamlessly facilitating cross-modal integration through unified architectures. Compared to conventional deep learning techniques, unimodal foundation models offer a more cost-effective and efficient solution for representation learning~\cite{xu2024framework}. Furthermore, the emergence of multimodal foundation models further extends their capabilities by enabling seamless integration of information across heterogeneous data sources~\cite{qiu2024application}.

Despite increasing evidence highlighting the potential of foundation models in processing clinical knowledge~\cite{singhal2023large,saab2024capabilities}, concerns regarding their real-world applicability persist~\cite{yang2023large,alsaad2024multimodal}. Given the high safety and ethical standards in clinical practice~\cite{yang2024disparities}, the deployment of such models must be approached cautiously, ensuring they are thoroughly validated as clinically beneficial, fair, and reliable before implementation~\cite{wornow2023shaky}. Additionally, existing benchmark studies~\cite{ye2024gmai,chen2023multimodal} have predominantly evaluated unimodal or multimodal foundation models separately, lacking a unified evaluation framework for systematic comparison. Consequently, a significant gap remains in understanding how well foundation models perform across diverse tasks and data modalities, underscoring the need for a comprehensive assessment of their capabilities in clinical settings.

\begin{figure*}[!t]
\centering
\includegraphics[width=0.85\textwidth]{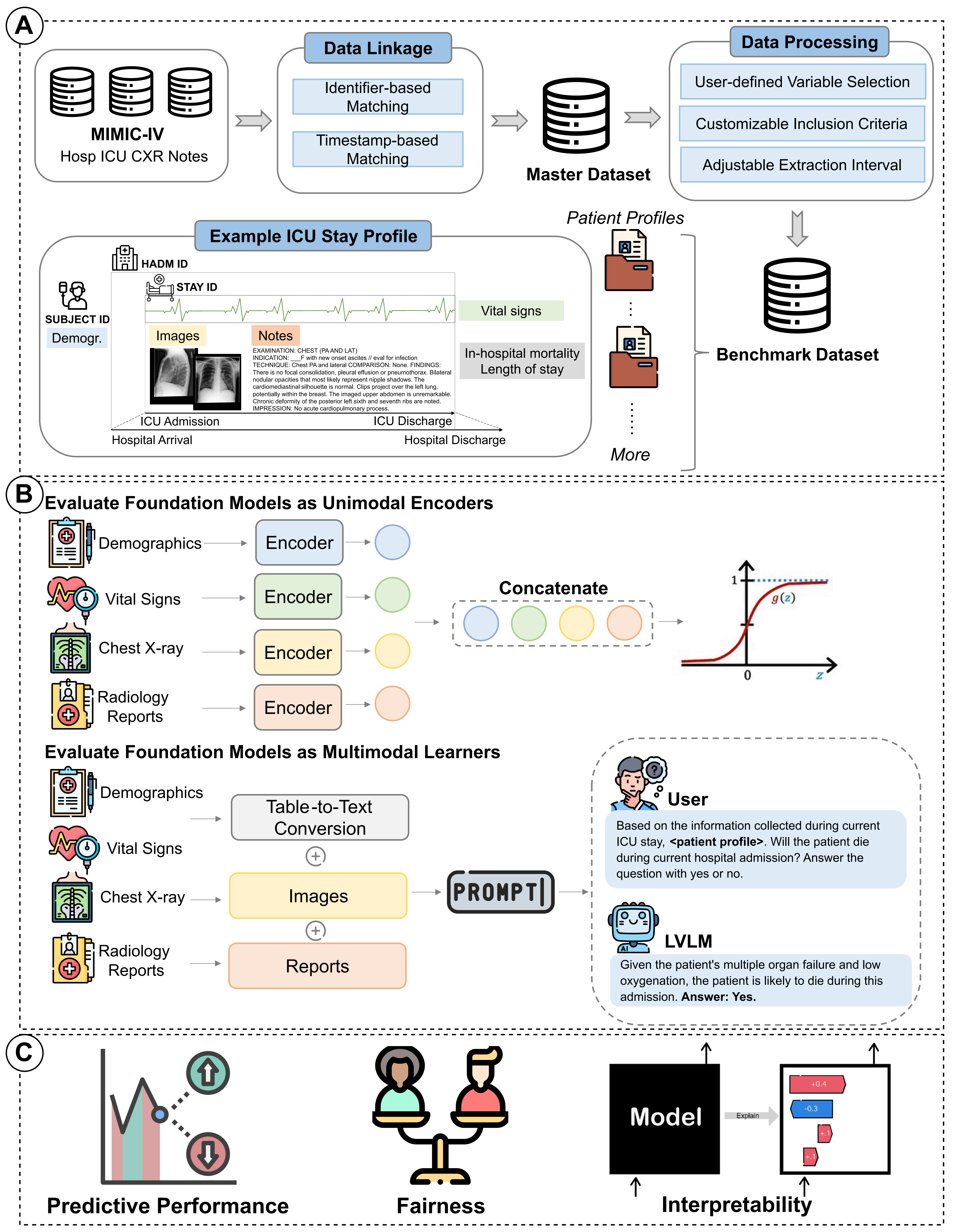}
\caption{An illustration of the three contributions of this study. (A) a flexible integration pipeline for standardized data; (B) a comprehensive benchmark for foundation models with multimodal data; and (C) an evaluation framework incorporating predictive performance, fairness, and interpretability assessments.}
\label{fig1}
\end{figure*}

To address the above concerns, this study aims to systematically evaluate the capabilities of foundation models in medical applications by using the publicly available Medical Information Mart for Intensive Care (MIMIC)-IV database~\cite{johnson2023mimic}. As illustrated in Figure~\ref{fig1}, we propose a holistic multimodal clinical benchmark encompassing data pre-processing, framework construction, and comprehensive evaluation. By offering this new benchmark, we seek to provide comprehensive insights into the capabilities of current foundation models in medicine and encourage their responsible integration into clinical practice. Our main contributions include: 
\begin{itemize}
    \item We developed a flexible data processing pipeline that unifies complex multimodal EHR data into a standardized format for efficient use in downstream analysis.
    \item We established a multidimensional benchmark to systematically evaluate foundation models both as unimodal encoders and as multimodal learners, assessing their performance across diverse medical tasks and multimodal data types.
    \item We conducted a rigorous evaluation that extends beyond traditional predictive performance, encompassing dimensions such as fairness and interpretability.
\end{itemize}

\section{Materials and Methods}
\subsection{Data integration pipeline}
\subsubsection{Master dataset generation}
{In this study, we utilized the MIMIC-IV v2.2 database, a public dataset comprising de-identified health records from patients admitted to the ICU at the Beth Israel Deaconess Medical Center in Boston, MA. Our analysis included data on 73,181 ICU stays involving 50,920 unique patients and 66,239 hospital admissions. We constructed a master dataset at the ICU stay level through key identifier and timestamp-based matching, ensuring that each data instance corresponds to a distinct ICU stay with its associated clinical information. Following the Holistic AI in Medicine (HAIM)~\cite{soenksen2022integrated} framework, we collected data across four core modules of the MIMIC-IV database: Hospital, ICU, Chest X-ray (CXR), and Notes. This comprehensive integration captured the majority of patient care activities recorded during hospital encounters, providing a rich and versatile data source for a wide range of downstream medical tasks.}
\subsubsection{Data processing}
{The heterogeneity of data recorded in our master dataset necessitated a data processing pipeline capable of generating standardized inputs to ensure both efficiency and reproducibility. In this study, we focused on four key data modalities relevant to ICU monitoring: structured patient demographics, and time-series vital signs, CXR images, and free-text clinical notes. Each modality was processed individually and subsequently integrated into a unified, analysis-ready structure, forming the basis of our benchmark dataset.

To enhance flexibility, the pipeline was designed to accommodate user-defined criteria. Automatic variable selection was performed based on a user-specified variable list. For dynamic data types, such as time-series measurements, images, and clinical notes, which may contain multiple time-stamped entries, the pipeline enabled extraction within specified timeframes relative to ICU admission, with all data sorted chronologically. In this study, we focused on the first 24 hours of ICU stay. 

Additionally, the pipeline supported customizable cohort selection criteria, including patient age, availability of specific data modalities, and minimum ICU stay duration. Specifically for this study, we included only adult patients who were 18 years of age or older at the time of admission.
}
\subsection{Benchmark tasks}
\subsubsection{In-hospital mortality}
{Our first benchmark task is predicting in-hospital mortality, a crucial outcome in ICU settings where patients are often critically ill. In-hospital mortality serves as a key indicator of patient prognosis and the quality of intensive care provided. Accurate prediction of this outcome is vital for identifying patients at the highest risk of clinical deterioration, thereby facilitating timely and targeted interventions that can significantly improve patient outcomes.}
\subsubsection{Length of stay (LOS)}
{The second benchmark task focuses on predicting the length of ICU stay, specifically, identifying stays that exceed 3 days. Early identification of patients likely to experience prolonged ICU stays allows clinicians to optimize care planning, improve patient flow management, and allocate medical resources more effectively.}

\subsection{Evaluating foundation models as unimodal encoders}
\begin{figure*}[!t]
\centering
\includegraphics[width=0.9\textwidth]{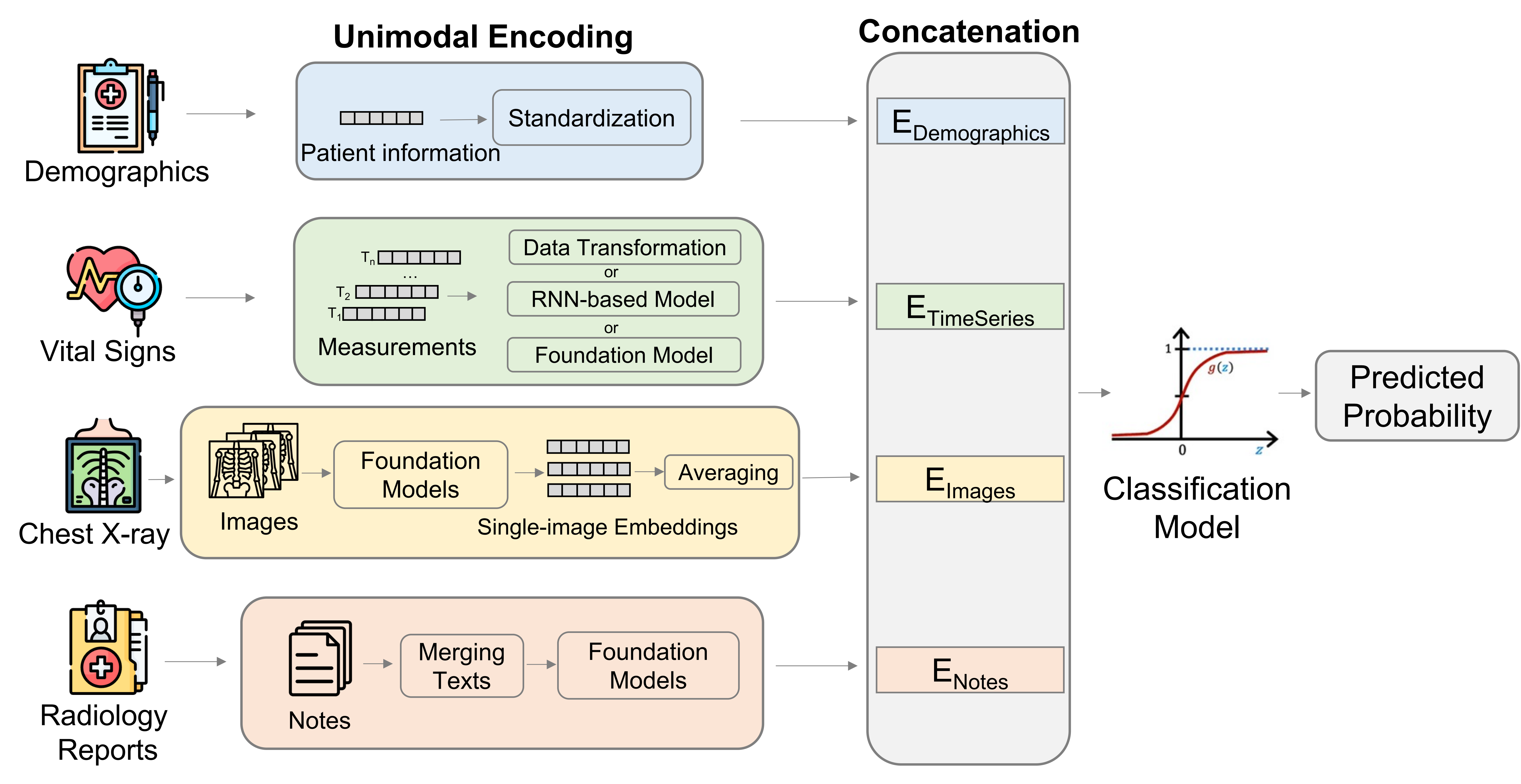}
\caption{Overview of the two-stage modular framework for evaluating foundation models as 
Unimodal encoders.}
\label{fig2}
\end{figure*}

Inspired by multimodal architectures described in prior studies~\cite{chen2023multimodal,soenksen2022integrated,shao2024multimodal}, we implemented a two-stage modular framework to assess the representation learning capabilities of foundation models, serving as unimodal encoders. An overview of the framework is shown in Figure~\ref{fig2}, with detailed embedding extraction approaches for each modality provided in Table~\ref{tab1}. In the first stage, embeddings for each data modality were extracted independently using the corresponding unimodal foundation models. In the second stage, these embeddings were concatenated to form a unified representation, which then served as the input for a logistic regression (LR) model used in downstream training and evaluation. This modular design enabled direct comparison of representation quality across unimodal encoders. All experiments were conducted using an 80/20 training and testing data split. 

\begin{table*}[!t]
\centering
\caption{Comparison of Embedding Techniques Across Modalities}
\begin{tabular}{lcccc}
\hline
\textbf{Techniques} & \textbf{Foundation model} & \textbf{Domain specificity} & \textbf{Training data size} & \textbf{Embedding dimension} \\
\hline
\multicolumn{5}{l}{\textbf{\textit{Time Series}}} \\

Fixed time interval   & No  & Domain-specific  & /                         & 312   \\
GRU~\cite{cho2014learning}                   & No  & Domain-specific  & 58,545 ICU stays          & 1,024 \\
Moment~\cite{goswami2024moment}  & Yes & General-purpose  & 13 million time series    & 1,024 \\
\hline
\multicolumn{5}{l}{\textbf{\textit{Image}}} \\
CXR-Foundation~\cite{xu2023elixr}     & Yes & Domain-specific  & Over 800,000 chest X-ray images & 1,376 \\
Swin Transformer~\cite{liu2021swin}  & Yes & General-purpose  & 14.2 million images               & 1,024 \\
\hline
\multicolumn{5}{l}{\textbf{\textit{Text}}} \\
RadBERT~\cite{chambon2023improved}        & Yes & Domain-specific  & 4,056,227 radiology reports                 & 768   \\
Text-Embedding-3-Large        & Yes & General-purpose  & Hundreds of billions of tokens    & 1,024 \\
\hline
\end{tabular}
\label{tab1}
\end{table*}

\subsubsection{Structured data}
{Structured data in this study included patient demographics and continuously measured vital signs. To ensure data quality, we removed outliers following the procedure outlined in~\cite{xie2022benchmarking} and addressed missing values through forward imputation. Numerical demographic variables were standardized, while categorical variables were encoded using one-hot encoding. For time-series vital signs, we evaluated three distinct embedding extraction methods. The first one involved fixed time interval aggregation, wherein irregularly sampled measurements were transformed into fixed-dimensional representations by averaging values over 1-hour intervals.The second method employed a gated recurrent unit (GRU) model~\cite{cho2014learning}, a deep learning approach designed to capture temporal dependencies in sequential data. The GRU model was specifically trained to generate embeddings from its final hidden state, using inputs that included vital signs, binary masking indicators for missing values, and relative time interval vectors. The third method utilized Moment~\cite{goswami2024moment}, a general-purpose time-series foundation model. Here, we applied Moment without domain-specific fine-tuning to assess its zero-shot performance on medical data.}

\subsubsection{Images}
{
For learning representations from CXR images, we evaluated two foundation models with varied domain specificity. The first, CXR-Foundation~\cite{xu2023elixr}, was designed specifically for clinical applications and trained on over 800,000 CXRs. In contrast, the Swin Transformer~\cite{liu2021swin} was pre-trained on the ImageNet-22K dataset with a hierarchical architecture to effectively capture both local and global visual features. For each ICU stay, chest X-rays were individually processed by the models, and the generated embeddings were averaged to obtain a unified image representation.
}

\subsubsection{Notes}
{
Similarly, radiology reports were encoded using both domain-specific and general-purpose models. The first model, RadBERT~\cite{chambon2023improved}, is a domain-adapted variant of BERT, fine-tuned on radiology reports to optimize its understanding of clinical terminology and contextual nuance. For the general-purpose model, we selected OpenAI’s text-embedding-3-large, known for its strong performance across diverse natural language processing (NLP) tasks. To preserve the sequential information, all reports within a given stay were combined in chronological order prior to embedding extraction.
}

\subsection{Evaluating foundation models as multimodal learners }
\begin{figure*}[!t]
\centering
\includegraphics[width=0.9\textwidth]{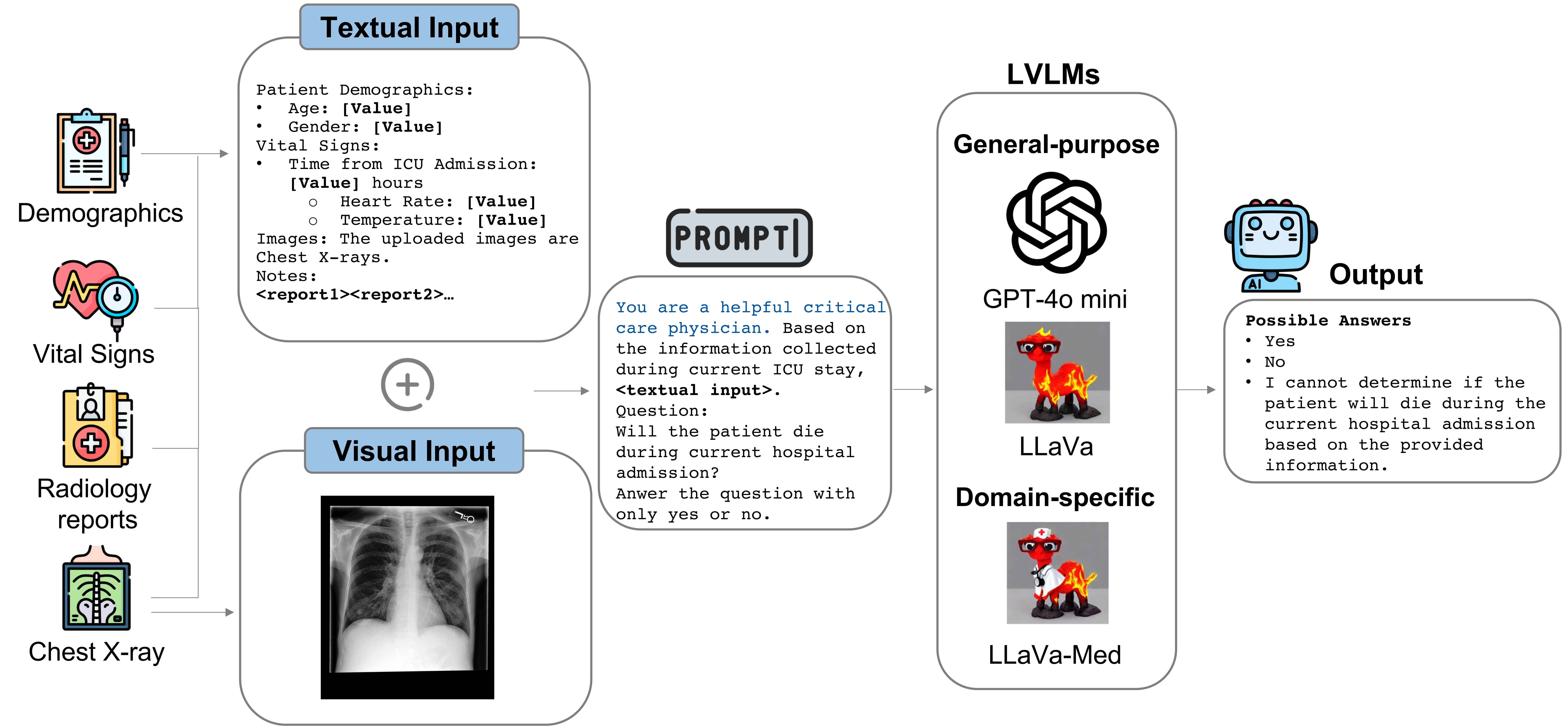}
\caption{Overview of the evaluation framework for evaluating foundation models as multimodal learners.}
\label{fig3}
\end{figure*}

Beyond unimodal representation learning, we further evaluated the multimodal reasoning capabilities of large vision-language models (LVLMs) across both general-purpose and medical domains. Specifically, we assessed LLaVA-Med~\cite{li2023llava}, a domain-specialized model tailored for clinical applications, alongside general-purpose models including GPT-4o mini and LLaVA-v1.5-7b~\cite{liu2023visual}. Each model was prompted with combined textual and visual inputs to predict both in-hospital mortality and length of stay. To ensure compliance with the MIMIC-IV data use policy, we accessed all OpenAI models through the Microsoft Azure OpenAI service. Figure~\ref{fig3} illustrates the overall benchmark workflow.

To standardize input formatting, we utilized GPT-4o to design a template converting non-image clinical data into a unified text input. The template performed a table-to-text conversion for demographic information and vital signs. Radiology reports were directly integrated as free-text inputs. The temporal structure of patient records was preserved by appending timestamps with their corresponding clinical records.

We adopted a closed-ended yes/no question format to systematically evaluate model performance on predicting patient outcomes. For example, in case of mortality prediction, each model was prompted with the following query:\textit{“Based on the information collected during current ICU stay, \textless patient profile\textgreater. Question: Will the patient die during current hospital admission? Answer the question using only yes or no.”}

\subsection{Evaluation}
\subsubsection{Predictive performance}
{
We adopted different evaluation metrics for benchmarking foundation models as unimodal encoders versus multimodal learners. For unimodal encoder evaluation, we reported the Area Under the Receiver Operating Characteristic Curve (AUROC), Area Under the Precision-Recall Curve (AUPRC), and accuracy. To ensure the robustness of performance estimates, we employed 1,000 bootstrapped samples to calculate 95\% confidence intervals for all metrics. For multimodal learner evaluation, we reported accuracy, precision, recall, specificity, and F1-score. To assess the models' refusal-to-answer behavior, we reported the percentage of answerable questions.
}

\subsubsection{Fairness}
{
Bias can emerge at various stages in the development of AI frameworks, potentially exacerbating existing inequities and undermining trust in clinical decision-making. To promote fairness awareness within our benchmark, we compared model performance across subgroups defined by sensitive variables such as age, gender, and race. We also employed two widely used group-based fairness metrics: demographic parity and equalized odds. Demographic parity evaluates whether model predictions are independent of sensitive attributes, while equalized odds measures the consistency of model performance, specifically true positive and false positive rates, across subgroups. Demographic parity was quantified as the ratio of the lowest to the highest selection rates among groups, and equalized odds was defined as the smaller value between the ratios of true positive rates and false positive rates across groups. Values closer to one reflect greater fairness.
}

\subsubsection{Interpretability}
{
To evaluate feature importance and model interpretability in clinical settings, we employed both SHapley Additive exPlanations (SHAP)~\cite{lundberg2017unified} and logistic regression coefficients in the modular framework. SHAP is a model-agnostic, game-theoretic approach that attributes predictions to individual features, offering both instance-level and global explanations. Logistic regression coefficients leverage the interpretability of linear models to directly quantify each feature's contribution to the final prediction.
}

\section{Results}
\subsection{Evaluating foundation models as unimodal encoders}
\subsubsection{Predictive performance}
{
Table~\ref{tab2} and~\ref{tab3} show our results, beginning with baseline models using only structured data. The GRU model outperformed both fixed-interval aggregation and Moment across both prediction tasks. We then evaluated the incremental value of incorporating image and text embeddings. Notably, for both modalities, domain-specific foundation models achieved performances comparable to general-purpose models pretrained at a much larger scale. Integrating multimodal data consistently improved predictive performance compared to models trained exclusively on structured data, suggesting that multimodal integration can enrich the information available for healthcare predictions and enhance accuracy. However, a modest decline in performance was observed when both image and text modalities were combined, compared to using text alone. This decline may stem from missing modalities or ineffective multimodal fusion strategies.
}

\begin{table*}[!t]
\centering
\begin{threeparttable}
\caption{Comparison of the performance of different models for in-hospital mortality prediction}
\label{tab2}
\begin{tabular}{p{3.2cm}>{\raggedright\arraybackslash}p{4cm}ccc}
\toprule
\textbf{Modalities} & \textbf{Embedding techniques} & \textbf{AUROC (95\% CI)} & \textbf{AUPRC (95\% CI)} & \textbf{Accuracy (95\% CI)} \\
\midrule

\multirow{3}{*}{\textbf{Structured Data (Baseline)}} 
& Fixed Time Interval & 0.8544 [0.8453, 0.8631] & 0.5187 [0.4943, 0.5417] & 0.8993 [0.8947, 0.9042] \\
& GRU & \underline{0.8778 [0.8700, 0.8865]} & \underline{0.5927 [0.5687, 0.6139]} & \underline{0.9101 [0.9054, 0.9145]} \\
& Moment & 0.7337 [0.7210, 0.7462] & 0.3197 [0.2972, 0.3435] & 0.8880 [0.8830, 0.8934] \\
\midrule

\multirow{2}{*}{\textbf{+Images}} 
& CXR Foundation & \underline{0.8801 [0.8720, 0.8878]} & \underline{0.5972 [0.5741, 0.6199]} & 0.9095 [0.9048, 0.9140] \\
& Swin Transformer & 0.8794 [0.8712, 0.8873] & 0.5960 [0.5742, 0.6193] & \underline{0.9105 [0.9060, 0.9155]} \\
\midrule

\multirow{2}{*}{\textbf{+Notes}} 
& RadBERT & 0.8876 [0.8800, 0.8944] & 0.6067 [0.5847, 0.6295] & 0.9091 [0.9049, 0.9139] \\
& Text-Embedding-3-Large & \underline{\textbf{0.8943 [0.8863, 0.9018]}} & \underline{\textbf{0.6262 [0.6029, 0.6486]}} & \underline{\textbf{0.9118 [0.9072, 0.9163]}} \\
\midrule

\multirow{6}{*}{\textbf{+Images \& Notes}} 
& CXR Foundation, RadBERT & 0.8886 [0.8806, 0.8962] & 0.6106 [0.5892, 0.6334] & \underline{0.9107 [0.9059, 0.9154]} \\
& CXR Foundation, Text-Embedding-3-Large & 0.8846 [0.8767, 0.8931] & 0.6035 [0.5812, 0.6269] & 0.9094 [0.9045, 0.9137] \\
& Swin Transformer, RadBERT & 0.8885 [0.8805, 0.8964] & 0.6096 [0.5883, 0.6320] & 0.9101 [0.9057, 0.9146] \\
& Swin Transformer, Text-Embedding-3-Large & \underline{0.8918 [0.8837, 0.8994]} & \underline{0.6187 [0.5972, 0.6410]} & 0.9106 [0.9057, 0.9149] \\
\bottomrule
\end{tabular}
\begin{tablenotes}
\scriptsize
\item[*] \underline{Underlined} values indicate the best performance within each modality group; \textbf{Bold} values indicate the best performance across all combinations.
\item[*] The best baseline was selected based on the highest AUROC among three embedding techniques, with modality addition performed on the best baseline to evaluate the impact of incorporating additional data modalities.
\end{tablenotes}
\end{threeparttable}
\end{table*}

\begin{table*}[!t]
\centering
\begin{threeparttable}
\caption{Comparison of the performance of different models for length of stay prediction}
\label{tab3}
\begin{tabular}{p{3.2cm}>{\raggedright\arraybackslash}p{4cm}ccc}
\toprule
\textbf{Modalities} & \textbf{Embedding techniques} & \textbf{AUROC (95\% CI)} & \textbf{AUPRC (95\% CI)} & \textbf{Accuracy (95\% CI)} \\
\midrule

\multirow{3}{*}{\textbf{Structured Data (Baseline)}} & Fixed Time Interval & 0.7573 [0.7486, 0.7660] & 0.6152 [0.6000, 0.6329] & 0.7605 [0.7539, 0.7674] \\
& GRU & \underline{0.8410 [0.8338, 0.8473]} & \underline{0.7102 [0.6971, 0.7245]} & \underline{0.7826 [0.7761, 0.7891]} \\
& Moment & 0.6964 [0.6877, 0.7050] & 0.5132 [0.4982, 0.5286] & 0.7003 [0.6932, 0.7075] \\
\midrule

\multirow{2}{*}{\textbf{+Images}} & CXR Foundation & \underline{0.8428 [0.8361, 0.8491]} & \underline{0.7147 [0.7015, 0.7276]} & \underline{0.7835 [0.7771, 0.7898]} \\
& Swin Transformer & 0.8418 [0.8351, 0.8478] & 0.7129 [0.6988, 0.7270] & 0.7830 [0.7762, 0.7894] \\
\midrule

\multirow{2}{*}{\textbf{+Notes}} & RadBERT & 0.8494 [0.8428, 0.8558] & 0.7227 [0.7093, 0.7376] & 0.7846 [0.7784, 0.7912] \\
& Text-Embedding-3-Large & \textbf{\underline{0.8532 [0.8470, 0.8599]}} & \textbf{\underline{0.7297 [0.7161, 0.7434]}} & \textbf{\underline{0.7894 [0.7827, 0.7957]}} \\
\midrule

\multirow{6}{*}{\textbf{+Images\&Notes}} & CXR Foundation, RadBERT & 0.8501 [0.8442, 0.8567] & 0.7249 [0.7119, 0.7374] & 0.7870 [0.7806, 0.7937] \\
& CXR Foundation, Text-Embedding-3-Large & 0.8481 [0.8418, 0.8543] & 0.7227 [0.7081, 0.7375] & 0.7836 [0.7769, 0.7896] \\
& Swin Transformer, RadBERT & 0.8492 [0.8429, 0.8556] & 0.7229 [0.7087, 0.7361] & 0.7857 [0.7793, 0.7923] \\
& Swin Transformer, Text-Embedding-3-Large & \underline{0.8502 [0.8442, 0.8563]} & \underline{0.7267 [0.7129, 0.7398]} & \underline{0.7882 [0.7814, 0.7944]} \\
\bottomrule
\end{tabular}
\begin{tablenotes}
\scriptsize
\item[*] \underline{Underlined} values indicate the best performance within each modality group; \textbf{Bold} values indicate the best performance across all combinations.
\item[*] The best baseline was selected based on the highest AUROC among three embedding techniques, with modality addition performed on the best baseline to evaluate the impact of incorporating additional data modalities.
\end{tablenotes}
\end{threeparttable}
\end{table*}

\subsubsection{Fairness}
\begin{figure*}[!t]
\centering

\begin{subfigure}[b]{0.95\textwidth}
    \includegraphics[width=0.95\textwidth]{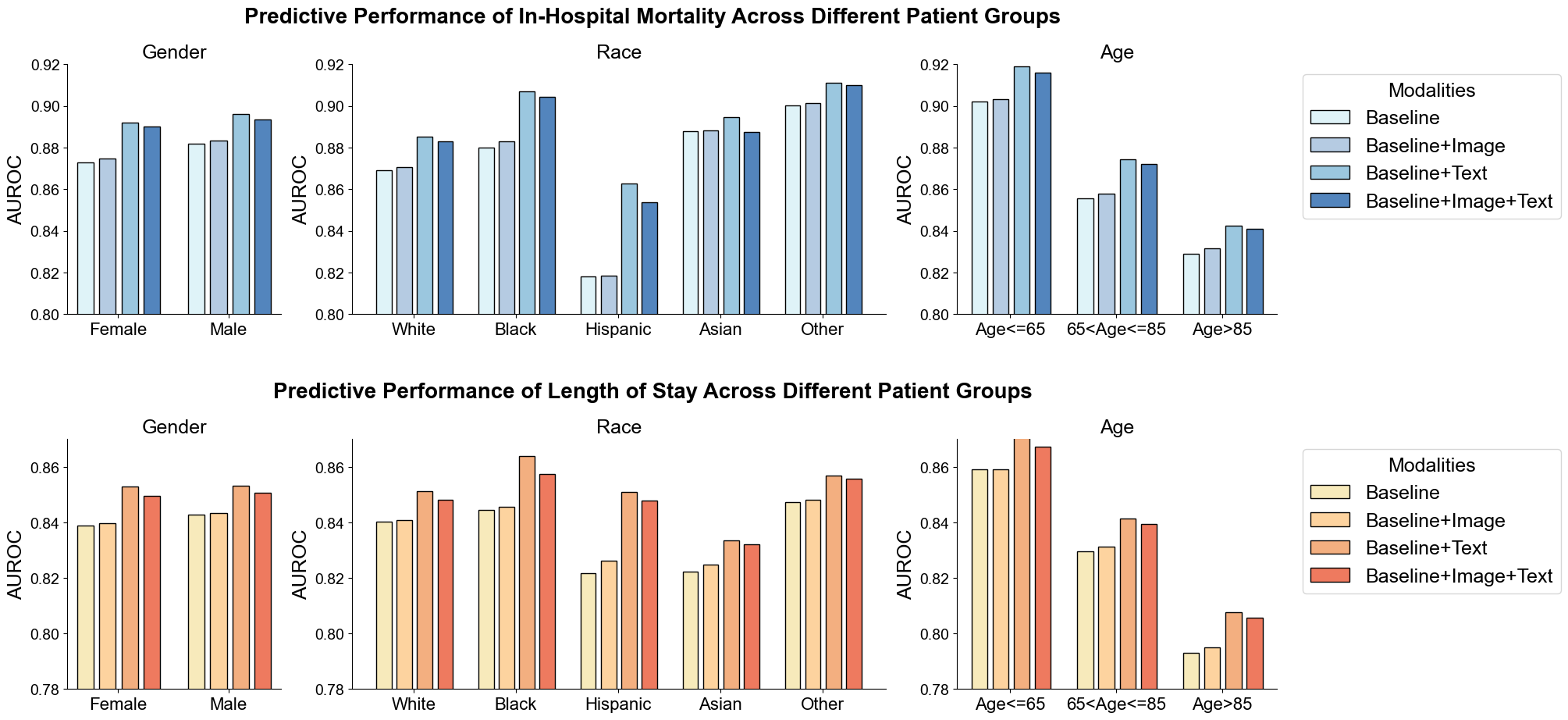}
    \caption{AUROC for in-hospital mortality and length-of-stay prediction stratified by gender, race, and age under different modality combinations.}
    \label{fig4A}
\end{subfigure}
\hfill
\begin{subfigure}[b]{0.95\textwidth}
    \includegraphics[width=0.95\textwidth]{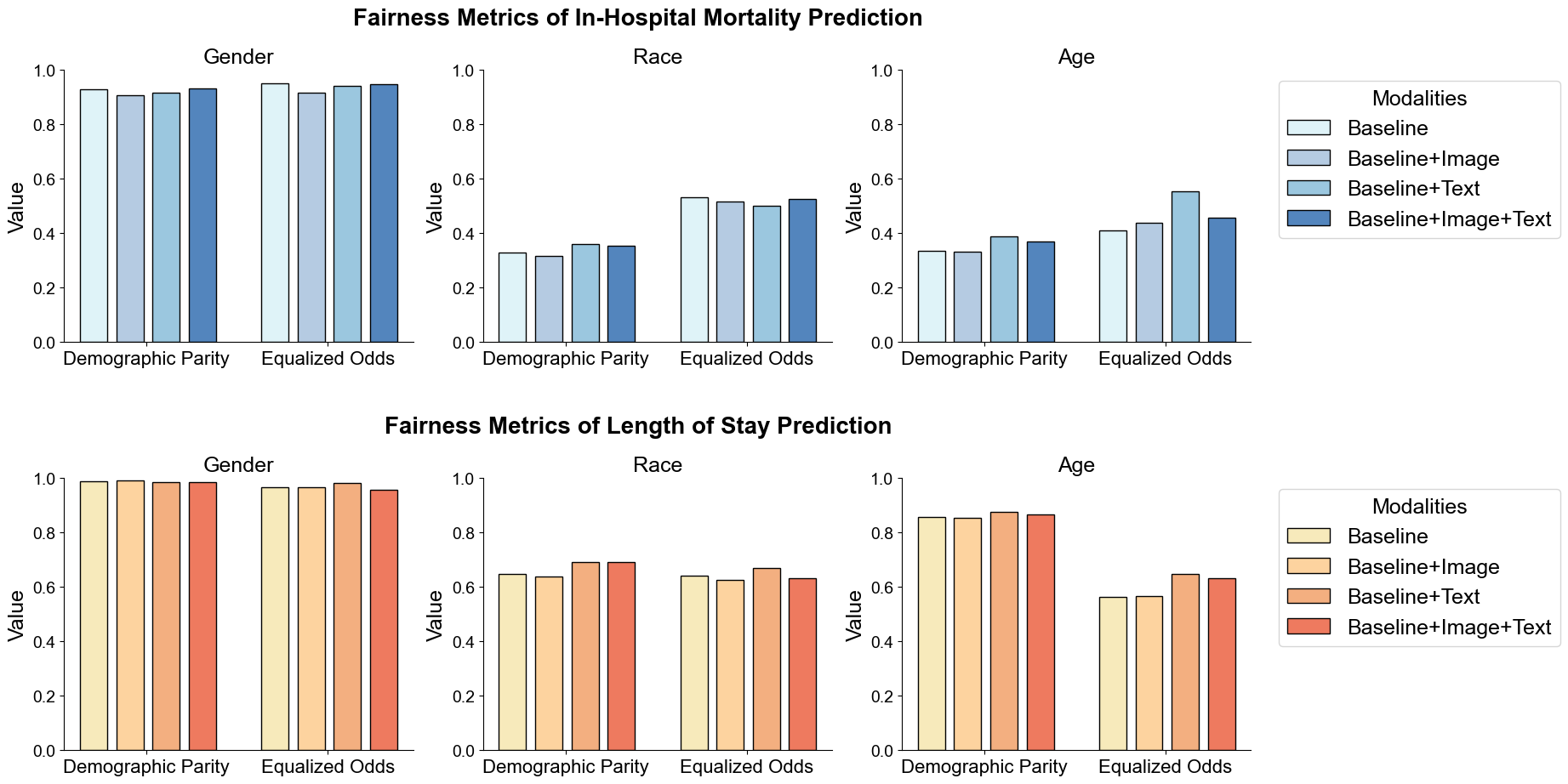}
    \caption{Fairness metrics under different modality combinations.}
    \label{fig4B}
\end{subfigure}

\caption{Subgroup analysis of predictive performance and fairness metrics across sensitive attributes.}
\label{fig4}
\end{figure*}

{
As shown in Figure~\ref{fig4}, model performance was relatively stable across gender subgroups but varied notably across race and age groups. In both prediction tasks, we observed a gradual decline in AUROC with increasing patient age, indicating reduced predictive accuracy in older populations. Racial disparities were also present, with lower performance for Hispanic patients in mortality prediction and for Asian patients in length-of-stay prediction. Despite these performance differences, the incorporation of multimodal data did not compromise fairness. As illustrated in Figure~\ref{fig4B}, demographic parity and equalized odds remained stable across subgroups, suggesting that the use of additional data modalities did not introduce further bias.
}

\subsubsection{Interpretability}
\begin{figure*}[!t]

\begin{subfigure}[b]{0.95\textwidth}
    \includegraphics[width=0.95\textwidth]{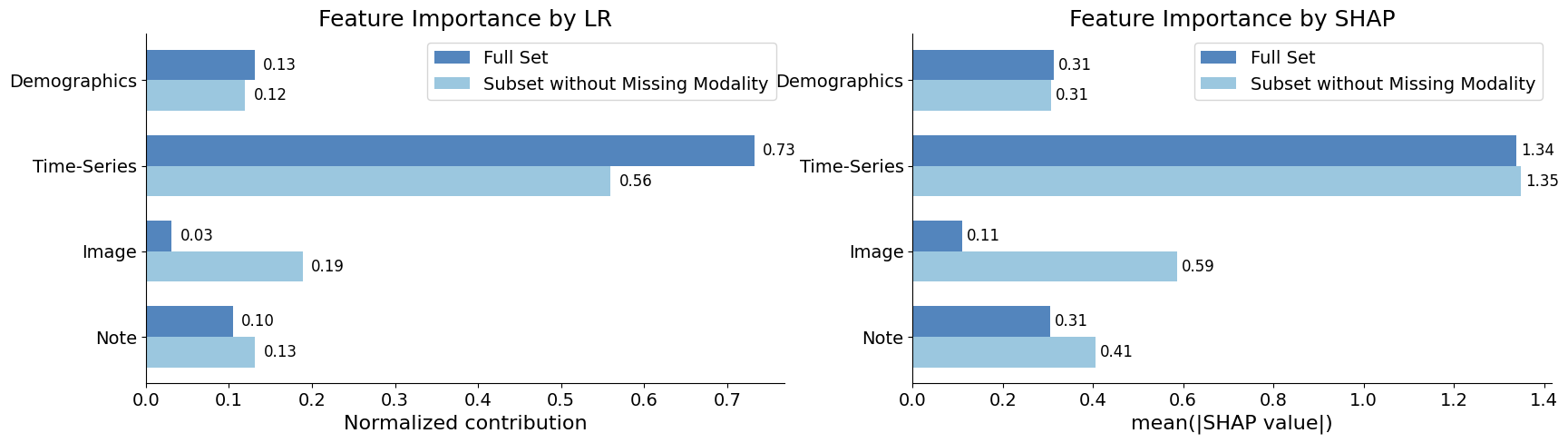}
    \caption{In-hospital mortality.}
    \label{fig5A}
\end{subfigure}
\hfill
\begin{subfigure}[b]{0.95\textwidth}
    \includegraphics[width=0.95\textwidth]{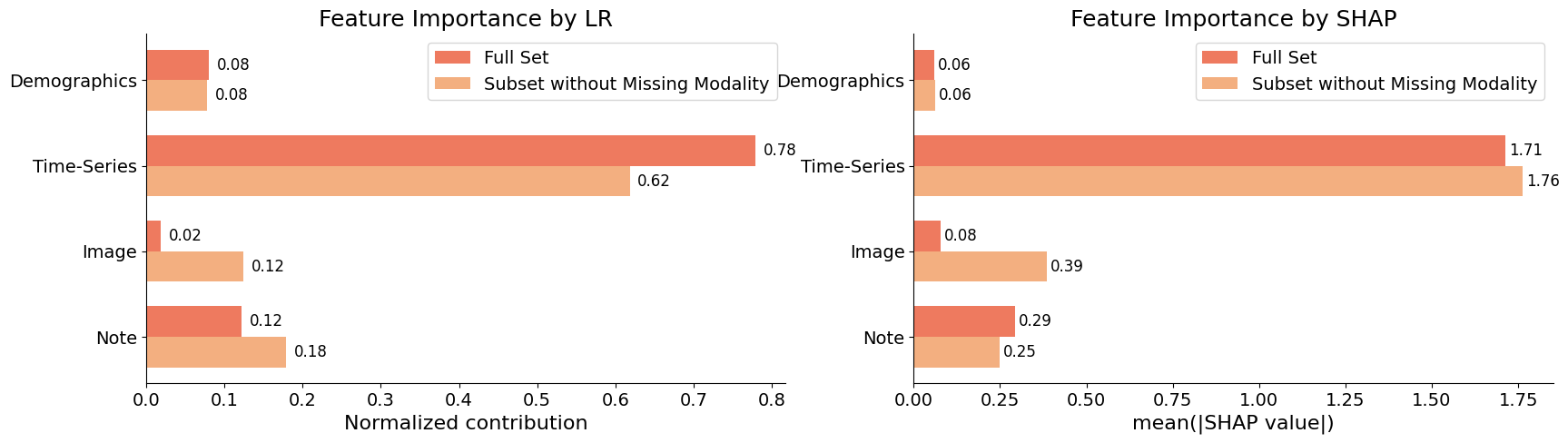}
    \caption{Length of stay.}
    \label{fig5B}
\end{subfigure}

\caption{Feature importance by modality using logistic regression coefficients and SHAP.}
\label{fig5}
\end{figure*}

{
Feature importance analysis, presented in Figure~\ref{fig5}, shows general consistency in the relative ranking of modality contributions. Time-series data emerged as the most influential modality across both prediction tasks. While imaging features had a limited impact in the full dataset, their contribution increased substantially in the subset of patients without missing modalities. These findings highlight the sensitivity of feature importance to data completeness and underscore the need to account for missing modalities in multimodal tasks.
}

\subsection{Evaluating foundation models as multimodal learners}
\subsubsection{Predictive performance}
\begin{figure}[!t]
\centerline{\includegraphics[width=\columnwidth]{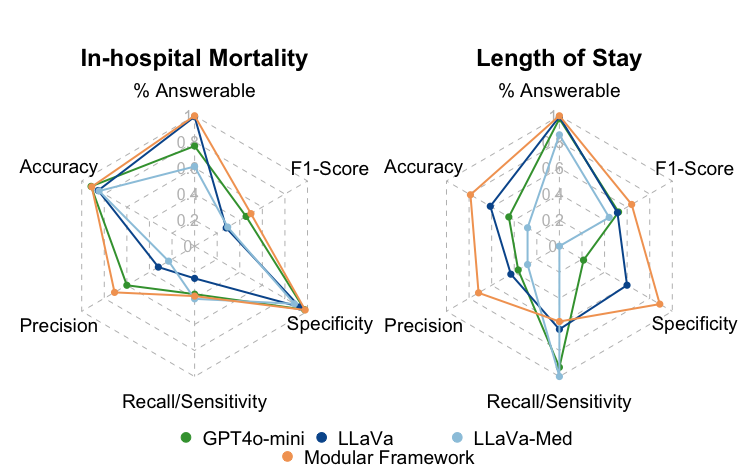}}
\caption{Comparison of predictive performance of MLLMs and modular framework.}
\label{fig6}
\end{figure}

{
As shown in Figure~\ref{fig6}, we conducted a comprehensive evaluation of multimodal large language models (MLLMs) across two prediction tasks. GPT4o-mini achieved performance comparable to the task-specific modular framework for in-hospital mortality prediction. However, all MLLMs demonstrated limited effectiveness in predicting length of stay. This highlights the current limitations in the task generalizability of LVLM. Notably, domain specialization did not guarantee improved performance, as LLaVA-Med failed to outperform general-purpose models in either task.
}

\subsubsection{Fairness}
\begin{figure*}[!t]
\centerline{\includegraphics[width=\textwidth]{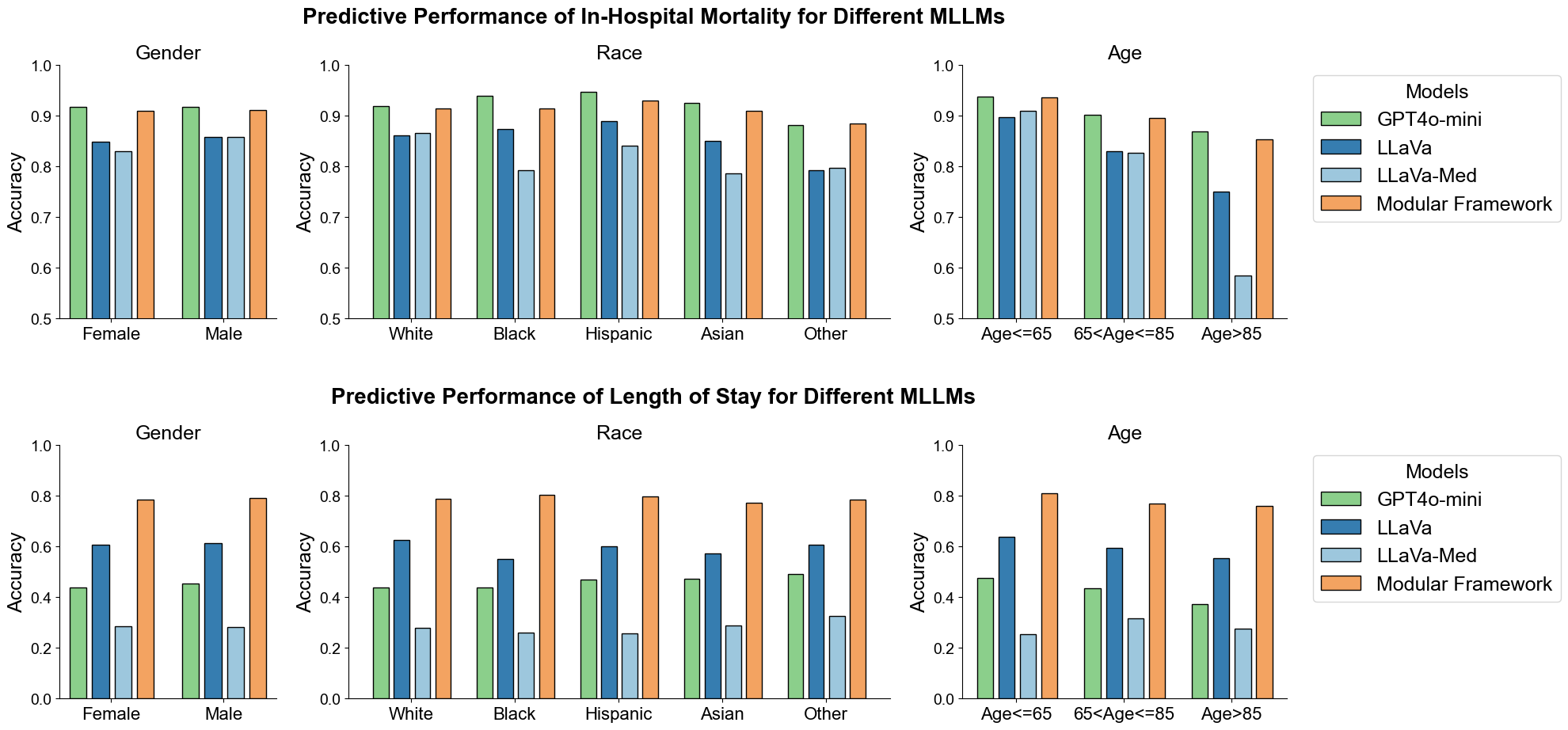}}
\caption{Subgroup analysis of predictive performance for MLLMs across sensitive attributes.}
\label{fig7}
\end{figure*}

{
Figure~\ref{fig7} presents subgroup performance for different MLLMs and the modular framework across sensitive variables. While model accuracy remained relatively consistent across gender for in-hospital mortality prediction, notable disparities were observed across racial and age groups. For length-of-stay prediction, MLLMs exhibited reduced performance across nearly all subgroups, highlighting substantial challenges in generalizing to this task across diverse populations.
}

\section{Discussion}
The emergence of foundation models presents a promising opportunity to advance clinical decision-making, yet their successful integration into real-world healthcare remains challenging. This gap is especially concerning given the high variability in data quality~\cite{sambasivan2021everyone}, the absence of unified benchmarks, and the critical need for trustworthy models~\cite{ning2024ethics}. Our work tackled these challenges by introducing a comprehensive benchmark built on multimodal public EHR data. A key strength of this study is the construction of a standardized data processing pipeline that harmonizes heterogeneous clinical data into consistent, analysis-ready inputs. By evaluating foundation models as both unimodal encoders and multimodal learners, we provided a holistic assessment of their ability to leverage diverse medical information for clinical prediction. Beyond predictive accuracy, we further evaluated fairness and feature importance to address growing concerns about the trustworthiness of AI in healthcare. Taken together, our benchmark serves as a critical step toward bridging the gap between model development and real-world deployment in healthcare.

Our benchmark provided valuable insights into the ways foundation models process multimodal medical data. First, we found that incorporating heterogeneous clinical data, which reflects real-world clinical practice, consistently enhanced predictive performance~\cite{topol2023artificial}. Second, while unimodal foundation models have achieved notable success in natural language processing~\cite{yang2024ascle} and computer vision, their capability to effectively process time-series data remains limited~\cite{gu2025time}. Third, our results indicate that domain-specific models, even when trained on a smaller scale, can achieve performance comparable to their larger counterparts, highlighting the cost-effectiveness of domain-adapted fine-tuning~\cite{hou2025enhancing}. Finally, although foundation models exhibit strong representation learning capabilities within individual data modalities, the adoption of effective multimodal fusion approaches plays a pivotal role in unlocking their full potential in addressing complex clinical tasks~\cite{chen2024artificial}.

In evaluating multimodal foundation models, we found that LVLMs showed promising performance in certain tasks, such as in-hospital mortality prediction, but struggled with others like length-of-stay prediction, indicating limited generalizability across clinical outcomes~\cite{liu2024visual}. This highlights current limitations of LVLMs in processing complex, context-rich clinical data~\cite{adams2024longhealth}. In practical applications, modular frameworks remain an effective alternative, as they leverage the strengths of foundation models while supporting task-specific adaptation. Notably, we observed that medical LVLMs such as LLaVA-Med showed inferior performance compared to general-purpose models~\cite{hu2024omnimedvqa}. These findings emphasize the need for continued efforts to develop more robust and versatile LVLMs that can effectively process multimodal medical data while integrating both medical knowledge and general reasoning capabilities~\cite{ye2024gmai}.

Beyond predictive performance, the trustworthiness of foundation models, such as fairness and interpretability, is indispensable for their deployment in healthcare~\cite{ning2024generative,xia2024cares}. Encouragingly, we observed that incorporating multiple data modalities did not increase bias, suggesting that performance gains from multimodal integration can be achieved without compromising fairness across patient subgroups~\cite{zhou2021radfusion}. However, handling missing modalities is a major challenge in multimodal EHR analysis due to inconsistent availability across patients~\cite{lee2023learning,zhang2022m3care}. Our interpretability analysis revealed that missing modalities could influence model behavior, often reducing the contribution of affected modalities and shifting overall feature importance patterns.

Despite these contributions, this study has several limitations. First, the fusion strategy employed relied on simple concatenation, which may not fully capture the complex interactions among different data modalities. Future work should investigate more sophisticated fusion techniques that better model inter-modality relationships and improve modality alignment~\cite{cui2023deep}. Second, although our processed dataset integrates key data types relevant to critical care, it does not include additional modalities available in MIMIC-IV, such as ICD codes and ECG signals, which could further enrich the dataset and support a broader range of clinical tasks~\cite{chen2023multimodal}. Nevertheless, the master dataset developed in this study provides a flexible and extensible resource for advancing clinical AI research. Lastly, our interpretability evaluation focused on modular frameworks and did not extend to MLLMs. Interpretability in MLLMs often relies on language-based explanations that differ from post-hoc feature attribution methods commonly used in traditional machine learning models~\cite{singh2024rethinking}. Recent studies have raised concerns about the faithfulness and reliability of these explanations, as MLLMs may generate plausible but ungrounded responses. This highlights the need for new interpretability evaluation protocols and techniques specifically designed for MLLMs to ensure transparency and reliability in clinical applications~\cite{dang2024explainable}.

\section*{References}
\vspace{-2em}
\bibliographystyle{IEEEtran}
\bibliography{main}

\begin{thebibliography}{10}
\providecommand{\url}[1]{#1}
\csname url@samestyle\endcsname
\providecommand{\newblock}{\relax}
\providecommand{\bibinfo}[2]{#2}
\providecommand{\BIBentrySTDinterwordspacing}{\spaceskip=0pt\relax}
\providecommand{\BIBentryALTinterwordstretchfactor}{4}
\providecommand{\BIBentryALTinterwordspacing}{\spaceskip=\fontdimen2\font plus
\BIBentryALTinterwordstretchfactor\fontdimen3\font minus \fontdimen4\font\relax}
\providecommand{\BIBforeignlanguage}[2]{{%
\expandafter\ifx\csname l@#1\endcsname\relax
\typeout{** WARNING: IEEEtran.bst: No hyphenation pattern has been}%
\typeout{** loaded for the language `#1'. Using the pattern for}%
\typeout{** the default language instead.}%
\else
\language=\csname l@#1\endcsname
\fi
#2}}
\providecommand{\BIBdecl}{\relax}
\BIBdecl

\bibitem{tang2024harnessing}
A.~S. Tang, S.~R. Woldemariam, S.~Miramontes, B.~Norgeot, T.~T. Oskotsky, and M.~Sirota, ``Harnessing ehr data for health research,'' \emph{Nature Medicine}, vol.~30, no.~7, pp. 1847--1855, 2024.

\bibitem{abul2019personalized}
N.~S. Abul-Husn and E.~E. Kenny, ``Personalized medicine and the power of electronic health records,'' \emph{Cell}, vol. 177, no.~1, pp. 58--69, 2019.

\bibitem{jensen2012mining}
P.~B. Jensen, L.~J. Jensen, and S.~Brunak, ``Mining electronic health records: towards better research applications and clinical care,'' \emph{Nature Reviews Genetics}, vol.~13, no.~6, pp. 395--405, 2012.

\bibitem{de2023guide}
J.~W. de~Kok, M.~{\'A}.~A. de~la Hoz, Y.~de~Jong, V.~Brokke, P.~W. Elbers, P.~Thoral, A.~Castillejo, T.~Trenor, J.~M. Castellano, A.~E. Bronchalo \emph{et~al.}, ``A guide to sharing open healthcare data under the general data protection regulation,'' \emph{Scientific data}, vol.~10, no.~1, p. 404, 2023.

\bibitem{johnson2023mimic}
A.~E. Johnson, L.~Bulgarelli, L.~Shen, A.~Gayles, A.~Shammout, S.~Horng, T.~J. Pollard, S.~Hao, B.~Moody, B.~Gow \emph{et~al.}, ``Mimic-iv, a freely accessible electronic health record dataset,'' \emph{Scientific data}, vol.~10, no.~1, p.~1, 2023.

\bibitem{ke2024comparing}
Y.~Ke, R.~Yang, and N.~Liu, ``Comparing open-access database and traditional intensive care studies using machine learning: bibliometric analysis study,'' \emph{Journal of Medical Internet Research}, vol.~26, p. e48330, 2024.

\bibitem{acosta2022multimodal}
J.~N. Acosta, G.~J. Falcone, P.~Rajpurkar, and E.~J. Topol, ``Multimodal biomedical ai,'' \emph{Nature medicine}, vol.~28, no.~9, pp. 1773--1784, 2022.

\bibitem{moor2023foundation}
M.~Moor, O.~Banerjee, Z.~S.~H. Abad, H.~M. Krumholz, J.~Leskovec, E.~J. Topol, and P.~Rajpurkar, ``Foundation models for generalist medical artificial intelligence,'' \emph{Nature}, vol. 616, no. 7956, pp. 259--265, 2023.

\bibitem{tu2024towards}
T.~Tu, S.~Azizi, D.~Driess, M.~Schaekermann, M.~Amin, P.-C. Chang, A.~Carroll, C.~Lau, R.~Tanno, I.~Ktena \emph{et~al.}, ``Towards generalist biomedical ai,'' \emph{Nejm Ai}, vol.~1, no.~3, p. AIoa2300138, 2024.

\bibitem{bommasani2021opportunities}
R.~Bommasani, D.~A. Hudson, E.~Adeli, R.~Altman, S.~Arora, S.~von Arx, M.~S. Bernstein, J.~Bohg, A.~Bosselut, E.~Brunskill \emph{et~al.}, ``On the opportunities and risks of foundation models,'' \emph{arXiv preprint arXiv:2108.07258}, 2021.

\bibitem{liu2022multimodal}
S.~Liu, X.~Wang, Y.~Hou, G.~Li, H.~Wang, H.~Xu, Y.~Xiang, and B.~Tang, ``Multimodal data matters: Language model pre-training over structured and unstructured electronic health records,'' \emph{IEEE Journal of Biomedical and Health Informatics}, vol.~27, no.~1, pp. 504--514, 2022.

\bibitem{li2024multimodal}
C.~Li, Z.~Gan, Z.~Yang, J.~Yang, L.~Li, L.~Wang, J.~Gao \emph{et~al.}, ``Multimodal foundation models: From specialists to general-purpose assistants,'' \emph{Foundations and Trends{\textregistered} in Computer Graphics and Vision}, vol.~16, no. 1-2, pp. 1--214, 2024.

\bibitem{he2024foundation}
Y.~He, F.~Huang, X.~Jiang, Y.~Nie, M.~Wang, J.~Wang, and H.~Chen, ``Foundation model for advancing healthcare: challenges, opportunities and future directions,'' \emph{IEEE Reviews in Biomedical Engineering}, 2024.

\bibitem{xu2024framework}
S.~Xu, H.~Gui, V.~Rotemberg, T.~Wang, Y.~T. Chen, and R.~Daneshjou, ``A framework for evaluating the efficacy of foundation embedding models in healthcare,'' \emph{medRxiv}, pp. 2024--04, 2024.

\bibitem{qiu2024application}
J.~Qiu, W.~Yuan, and K.~Lam, ``The application of multimodal large language models in medicine,'' \emph{The Lancet Regional Health--Western Pacific}, vol.~45, 2024.

\bibitem{singhal2023large}
K.~Singhal, S.~Azizi, T.~Tu, S.~S. Mahdavi, J.~Wei, H.~W. Chung, N.~Scales, A.~Tanwani, H.~Cole-Lewis, S.~Pfohl \emph{et~al.}, ``Large language models encode clinical knowledge,'' \emph{Nature}, vol. 620, no. 7972, pp. 172--180, 2023.

\bibitem{saab2024capabilities}
K.~Saab, T.~Tu, W.-H. Weng, R.~Tanno, D.~Stutz, E.~Wulczyn, F.~Zhang, T.~Strother, C.~Park, E.~Vedadi \emph{et~al.}, ``Capabilities of gemini models in medicine,'' \emph{arXiv preprint arXiv:2404.18416}, 2024.

\bibitem{yang2023large}
R.~Yang, T.~F. Tan, W.~Lu, A.~J. Thirunavukarasu, D.~S.~W. Ting, and N.~Liu, ``Large language models in health care: Development, applications, and challenges,'' \emph{Health Care Science}, vol.~2, no.~4, pp. 255--263, 2023.

\bibitem{alsaad2024multimodal}
R.~AlSaad, A.~Abd-Alrazaq, S.~Boughorbel, A.~Ahmed, M.-A. Renault, R.~Damseh, and J.~Sheikh, ``Multimodal large language models in health care: applications, challenges, and future outlook,'' \emph{Journal of medical Internet research}, vol.~26, p. e59505, 2024.

\bibitem{yang2024disparities}
R.~Yang, S.~V. Nair, Y.~Ke, D.~D’Agostino, M.~Liu, Y.~Ning, and N.~Liu, ``Disparities in clinical studies of ai enabled applications from a global perspective,'' \emph{NPJ digital medicine}, vol.~7, no.~1, p. 209, 2024.

\bibitem{wornow2023shaky}
M.~Wornow, Y.~Xu, R.~Thapa, B.~Patel, E.~Steinberg, S.~Fleming, M.~A. Pfeffer, J.~Fries, and N.~H. Shah, ``The shaky foundations of large language models and foundation models for electronic health records,'' \emph{npj digital medicine}, vol.~6, no.~1, p. 135, 2023.

\bibitem{ye2024gmai}
J.~Ye, G.~Wang, Y.~Li, Z.~Deng, W.~Li, T.~Li, H.~Duan, Z.~Huang, Y.~Su, B.~Wang \emph{et~al.}, ``Gmai-mmbench: A comprehensive multimodal evaluation benchmark towards general medical ai,'' \emph{Advances in Neural Information Processing Systems}, vol.~37, pp. 94\,327--94\,427, 2024.

\bibitem{chen2023multimodal}
E.~Chen, A.~Kansal, J.~Chen, B.~T. Jin, J.~Reisler, D.~E. Kim, and P.~Rajpurkar, ``Multimodal clinical benchmark for emergency care (mc-bec): A comprehensive benchmark for evaluating foundation models in emergency medicine,'' \emph{Advances in Neural Information Processing Systems}, vol.~36, pp. 45\,794--45\,811, 2023.

\bibitem{soenksen2022integrated}
L.~R. Soenksen, Y.~Ma, C.~Zeng, L.~Boussioux, K.~Villalobos~Carballo, L.~Na, H.~M. Wiberg, M.~L. Li, I.~Fuentes, and D.~Bertsimas, ``Integrated multimodal artificial intelligence framework for healthcare applications,'' \emph{NPJ digital medicine}, vol.~5, no.~1, p. 149, 2022.

\bibitem{shao2024multimodal}
J.~Shao, J.~Ma, Y.~Yu, S.~Zhang, W.~Wang, W.~Li, and C.~Wang, ``A multimodal integration pipeline for accurate diagnosis, pathogen identification, and prognosis prediction of pulmonary infections,'' \emph{The Innovation}, vol.~5, no.~4, 2024.

\bibitem{cho2014learning}
K.~Cho, B.~Van~Merri{\"e}nboer, C.~Gulcehre, D.~Bahdanau, F.~Bougares, H.~Schwenk, and Y.~Bengio, ``Learning phrase representations using rnn encoder-decoder for statistical machine translation,'' \emph{arXiv preprint arXiv:1406.1078}, 2014.

\bibitem{goswami2024moment}
M.~Goswami, K.~Szafer, A.~Choudhry, Y.~Cai, S.~Li, and A.~Dubrawski, ``Moment: A family of open time-series foundation models,'' \emph{arXiv preprint arXiv:2402.03885}, 2024.

\bibitem{xu2023elixr}
S.~Xu, L.~Yang, C.~Kelly, M.~Sieniek, T.~Kohlberger, M.~Ma, W.-H. Weng, A.~Kiraly, S.~Kazemzadeh, Z.~Melamed \emph{et~al.}, ``Elixr: Towards a general purpose x-ray artificial intelligence system through alignment of large language models and radiology vision encoders,'' \emph{arXiv preprint arXiv:2308.01317}, 2023.

\bibitem{liu2021swin}
Z.~Liu, Y.~Lin, Y.~Cao, H.~Hu, Y.~Wei, Z.~Zhang, S.~Lin, and B.~Guo, ``Swin transformer: Hierarchical vision transformer using shifted windows,'' in \emph{Proceedings of the IEEE/CVF international conference on computer vision}, 2021, pp. 10\,012--10\,022.

\bibitem{chambon2023improved}
P.~Chambon, T.~S. Cook, and C.~P. Langlotz, ``Improved fine-tuning of in-domain transformer model for inferring covid-19 presence in multi-institutional radiology reports,'' \emph{Journal of Digital Imaging}, vol.~36, no.~1, pp. 164--177, 2023.

\bibitem{xie2022benchmarking}
F.~Xie, J.~Zhou, J.~W. Lee, M.~Tan, S.~Li, L.~S. Rajnthern, M.~L. Chee, B.~Chakraborty, A.-K.~I. Wong, A.~Dagan \emph{et~al.}, ``Benchmarking emergency department prediction models with machine learning and public electronic health records,'' \emph{Scientific Data}, vol.~9, no.~1, p. 658, 2022.

\bibitem{li2023llava}
C.~Li, C.~Wong, S.~Zhang, N.~Usuyama, H.~Liu, J.~Yang, T.~Naumann, H.~Poon, and J.~Gao, ``Llava-med: Training a large language-and-vision assistant for biomedicine in one day,'' \emph{Advances in Neural Information Processing Systems}, vol.~36, pp. 28\,541--28\,564, 2023.

\bibitem{liu2023visual}
H.~Liu, C.~Li, Q.~Wu, and Y.~J. Lee, ``Visual instruction tuning,'' \emph{Advances in neural information processing systems}, vol.~36, pp. 34\,892--34\,916, 2023.

\bibitem{lundberg2017unified}
S.~M. Lundberg and S.-I. Lee, ``A unified approach to interpreting model predictions,'' \emph{Advances in neural information processing systems}, vol.~30, 2017.

\bibitem{sambasivan2021everyone}
N.~Sambasivan, S.~Kapania, H.~Highfill, D.~Akrong, P.~Paritosh, and L.~M. Aroyo, ``“everyone wants to do the model work, not the data work”: Data cascades in high-stakes ai,'' in \emph{proceedings of the 2021 CHI Conference on Human Factors in Computing Systems}, 2021, pp. 1--15.

\bibitem{ning2024ethics}
Y.~Ning, X.~Liu, G.~S. Collins, K.~G. Moons, M.~McCradden, D.~S.~W. Ting, J.~C.~L. Ong, B.~A. Goldstein, S.~K. Wagner, P.~A. Keane \emph{et~al.}, ``An ethics assessment tool for artificial intelligence implementation in healthcare: Care-ai,'' \emph{Nature medicine}, pp. 1--2, 2024.

\bibitem{topol2023artificial}
E.~J. Topol, ``As artificial intelligence goes multimodal, medical applications multiply,'' \emph{Science}, vol. 381, no. 6663, p. eadk6139, 2023.

\bibitem{yang2024ascle}
R.~Yang, Q.~Zeng, K.~You, Y.~Qiao, L.~Huang, C.-C. Hsieh, B.~Rosand, J.~Goldwasser, A.~Dave, T.~Keenan \emph{et~al.}, ``Ascle—a python natural language processing toolkit for medical text generation: development and evaluation study,'' \emph{Journal of Medical Internet Research}, vol.~26, p. e60601, 2024.

\bibitem{gu2025time}
X.~Gu, Y.~Liu, Z.~Mohsin, J.~Bedford, A.~Thakur, P.~Watkinson, L.~Clifton, T.~Zhu, and D.~Clifton, ``Are time series foundation models ready for vital sign forecasting in healthcare?'' in \emph{Machine Learning for Health (ML4H)}.\hskip 1em plus 0.5em minus 0.4em\relax PMLR, 2025, pp. 401--419.

\bibitem{hou2025enhancing}
Z.~Hou, H.~Liu, J.~Bian, X.~He, and Y.~Zhuang, ``Enhancing medical coding efficiency through domain-specific fine-tuned large language models,'' \emph{npj Health Systems}, vol.~2, no.~1, p.~14, 2025.

\bibitem{chen2024artificial}
X.~Chen, H.~Xie, X.~Tao, F.~L. Wang, M.~Leng, and B.~Lei, ``Artificial intelligence and multimodal data fusion for smart healthcare: topic modeling and bibliometrics,'' \emph{Artificial Intelligence Review}, vol.~57, no.~4, p.~91, 2024.

\bibitem{liu2024visual}
C.~Liu, Y.~Jin, Z.~Guan, T.~Li, Y.~Qin, B.~Qian, Z.~Jiang, Y.~Wu, X.~Wang, Y.~F. Zheng \emph{et~al.}, ``Visual--language foundation models in medicine,'' \emph{The Visual Computer}, pp. 1--20, 2024.

\bibitem{adams2024longhealth}
L.~Adams, F.~Busch, T.~Han, J.-B. Excoffier, M.~Ortala, A.~L{\"o}ser, H.~J. Aerts, J.~N. Kather, D.~Truhn, and K.~Bressem, ``Longhealth: A question answering benchmark with long clinical documents,'' \emph{arXiv preprint arXiv:2401.14490}, 2024.

\bibitem{hu2024omnimedvqa}
Y.~Hu, T.~Li, Q.~Lu, W.~Shao, J.~He, Y.~Qiao, and P.~Luo, ``Omnimedvqa: A new large-scale comprehensive evaluation benchmark for medical lvlm,'' in \emph{Proceedings of the IEEE/CVF Conference on Computer Vision and Pattern Recognition}, 2024, pp. 22\,170--22\,183.

\bibitem{ning2024generative}
Y.~Ning, S.~Teixayavong, Y.~Shang, J.~Savulescu, V.~Nagaraj, D.~Miao, M.~Mertens, D.~S.~W. Ting, J.~C.~L. Ong, M.~Liu \emph{et~al.}, ``Generative artificial intelligence and ethical considerations in health care: a scoping review and ethics checklist,'' \emph{The Lancet Digital Health}, vol.~6, no.~11, pp. e848--e856, 2024.

\bibitem{xia2024cares}
P.~Xia, Z.~Chen, J.~Tian, Y.~Gong, R.~Hou, Y.~Xu, Z.~Wu, Z.~Fan, Y.~Zhou, K.~Zhu \emph{et~al.}, ``Cares: A comprehensive benchmark of trustworthiness in medical vision language models,'' \emph{Advances in Neural Information Processing Systems}, vol.~37, pp. 140\,334--140\,365, 2024.

\bibitem{zhou2021radfusion}
Y.~Zhou, S.-C. Huang, J.~A. Fries, A.~Youssef, T.~J. Amrhein, M.~Chang, I.~Banerjee, D.~Rubin, L.~Xing, N.~Shah \emph{et~al.}, ``Radfusion: Benchmarking performance and fairness for multimodal pulmonary embolism detection from ct and ehr,'' \emph{arXiv preprint arXiv:2111.11665}, 2021.

\bibitem{lee2023learning}
K.~Lee, S.~Lee, S.~Hahn, H.~Hyun, E.~Choi, B.~Ahn, and J.~Lee, ``Learning missing modal electronic health records with unified multi-modal data embedding and modality-aware attention,'' in \emph{Machine Learning for Healthcare Conference}.\hskip 1em plus 0.5em minus 0.4em\relax PMLR, 2023, pp. 423--442.

\bibitem{zhang2022m3care}
C.~Zhang, X.~Chu, L.~Ma, Y.~Zhu, Y.~Wang, J.~Wang, and J.~Zhao, ``M3care: Learning with missing modalities in multimodal healthcare data,'' in \emph{Proceedings of the 28th ACM SIGKDD conference on knowledge discovery and data mining}, 2022, pp. 2418--2428.

\bibitem{cui2023deep}
C.~Cui, H.~Yang, Y.~Wang, S.~Zhao, Z.~Asad, L.~A. Coburn, K.~T. Wilson, B.~A. Landman, and Y.~Huo, ``Deep multimodal fusion of image and non-image data in disease diagnosis and prognosis: a review,'' \emph{Progress in Biomedical Engineering}, vol.~5, no.~2, p. 022001, 2023.

\bibitem{singh2024rethinking}
C.~Singh, J.~P. Inala, M.~Galley, R.~Caruana, and J.~Gao, ``Rethinking interpretability in the era of large language models,'' \emph{arXiv preprint arXiv:2402.01761}, 2024.

\bibitem{dang2024explainable}
Y.~Dang, K.~Huang, J.~Huo, Y.~Yan, S.~Huang, D.~Liu, M.~Gao, J.~Zhang, C.~Qian, K.~Wang \emph{et~al.}, ``Explainable and interpretable multimodal large language models: A comprehensive survey,'' \emph{arXiv preprint arXiv:2412.02104}, 2024.

\end{thebibliography}

\end{document}